\documentclass[runningheads,a4paper]{llncs}
\usepackage{amsmath,graphicx,url,siunitx}
\usepackage{adjustbox}
\usepackage{wrapfig}
\usepackage{multirow}
\usepackage{amssymb}
\usepackage{makecell}
\usepackage{color, colortbl}
\definecolor{Gray}{gray}{0.9}
\definecolor{citecolor}{RGB}{34,139,34}
\usepackage[breaklinks=true,colorlinks,citecolor=citecolor,bookmarks=false]{hyperref}
\urldef{\mailsa}\path|{jaylee, khpaeng}@lunit.io|

\begin{document}

\title{A Robust and Effective Approach Towards Accurate Metastasis Detection and pN-stage Classification in Breast Cancer}
\titlerunning{A Robust and Effective Approach for pN-stage Classification}  
%
\author{Byungjae Lee \and Kyunghyun Paeng}
\authorrunning{Byungjae Lee \and Kyunghyun Paeng} 
%
%
\institute{Lunit inc., Seoul, South Korea \\
\mailsa }

\maketitle              

\begin{abstract}
Predicting TNM stage is the major determinant of breast cancer prognosis and treatment. The essential part of TNM stage classification is whether the cancer has metastasized to the regional lymph nodes (N-stage). Pathologic N-stage (pN-stage) is commonly performed by pathologists detecting metastasis in histological slides. However, this diagnostic procedure is prone to misinterpretation and would normally require extensive time by pathologists because of the sheer volume of data that needs a thorough review. Automated detection of lymph node metastasis and pN-stage prediction has a great potential to reduce their workload and help the pathologist. Recent advances in convolutional neural networks (CNN) have shown significant improvements in histological slide analysis, but accuracy is not optimized because of the difficulty in the handling of gigapixel images. In this paper, we propose a robust method for metastasis detection and pN-stage classification in breast cancer from multiple gigapixel pathology images in an effective way. pN-stage is predicted by combining patch-level CNN based metastasis detector and slide-level lymph node classifier. The proposed framework achieves a state-of-the-art quadratic weighted kappa score of 0.9203 on the Camelyon17 dataset, outperforming the previous winning method of the Camelyon17 challenge. 

\end{abstract}
\section{Introduction}
\label{sec:intro}

When cancer is first diagnosed, the first and most important step is staging of the cancer by using the TNM staging system~\cite{sobin2011tnm}, the most commonly used system. Invasion to lymph nodes, highly predictive of recurrence~\cite{saadatmand2015influence}, is evaluated by pathologists (pN-stage) via detection of tumor lesions in lymph node histology slides from a surgically resected tissue. This diagnostic procedure is prone to misinterpretation and would normally require extensive time by pathologists because of the sheer volume of data that needs a thorough review. Automated detection of lymph node metastasis and pN-stage prediction has the potential to significantly elevate the efficiency and diagnostic accuracy of pathologists for one of the most critical diagnostic process of breast cancer.

In the last few years, considerable improvements have been emerged in the computer vision task using CNN~\cite{he2016deep}. Followed by this paradigm, CNN based computer assisted metastasis detection has been proposed in recent years~\cite{paeng2017unified,liu2017detecting,bejnordi2017diagnostic}. However, recent approaches metastasis detection in whole slide images have shown the difficulty in handling gigapixel images~\cite{paeng2017unified,liu2017detecting,bejnordi2017diagnostic}. Furthermore, pN-stage classification requires handling multiple gigapixel images.

In this paper, we introduce a robust method to predict pathologic N-stage (pN-stage) from whole slide pathology images. For the robust performance, we effectively handle multiple gigapixel images in order to integrate CNN into pN-stage prediction framework such as balanced patch sampling, patch augmentation, stain color augmentation, 2-stage fine-tuning and overlap tiling strategy. We achieved patient-level quadratic weighted kappa score 0.9203 on the Camelyon17 \texttt{test} set which it yields the new state-of-the-art record on Camelyon17 leaderboard~\cite{camelyon17}. 

\begin{figure*}
 \vspace{-0.5 cm}
 \center
 \includegraphics[width=0.9\textwidth]{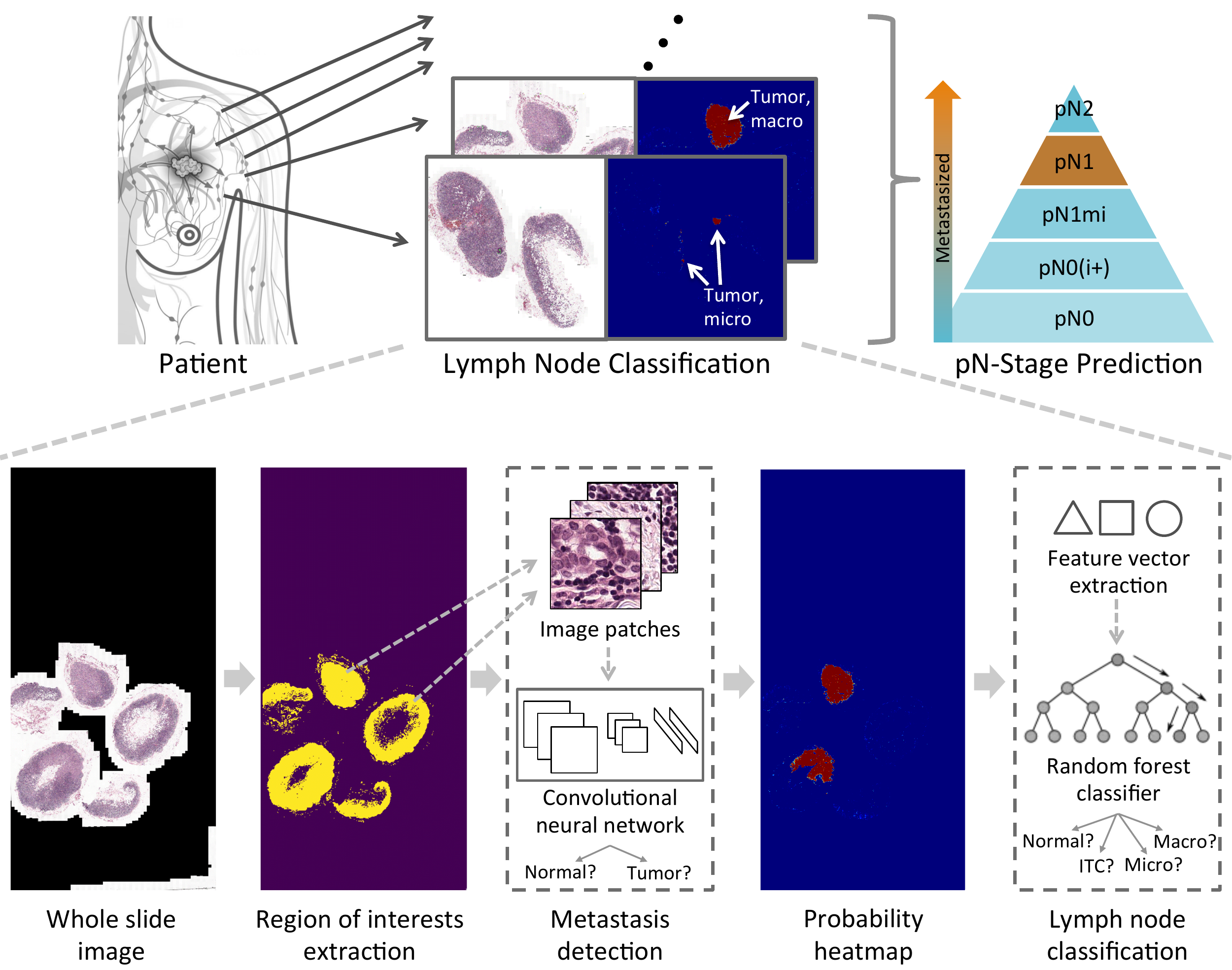}
 \caption{Overall architecture of our pN-stage prediction framework.}
 \label{fig1}
 \vspace{-0.7 cm}
\end{figure*}

\section{Methodology}
\label{sec:method}

Fig.~\ref{fig1} shows the overall scheme of our proposed framework. First, ROI extraction module proposes candidate tissue regions from whole slide images. Second, CNN-based metastasis detection module predicts cancer metastasis within extracted ROIs. Third, the predicted scores extracted from ROI are converted to a feature vector based on the morphological and geometrical information which is used to build a slide-level lymph node classifier. Patient-level pN-stage is determined by aggregating slide-level predictions with given rules~\cite{camelyon17}. 

\subsection{Regions of Interests Extraction}
\label{ssec:roi}

A whole slide image (WSI) is approximately 200000$\times$100000 pixels on the highest resolution level. Accurate tissue region extraction algorithms can save computation time and reduce false positives from noisy background area. In order to extract tissue regions from the WSIs, Otsu threshold~\cite{otsu1979threshold} or gray value threshold is commonly used in recent studies~\cite{paeng2017unified,liu2017detecting,bejnordi2017diagnostic}. We decide to use gray value threshold method which shows superior performance in our experiments. 

\subsection{Metastasis Detection}
\label{ssec:cnn}

Some annotated metastasis regions include non-metastasis area since accurate pixel-level annotation is difficult in gigapixel WSIs~\cite{liu2017detecting}. We build a large scale dataset by extracting small patches from WSIs to deal with those noisy labels. After the ROIs are found from WSIs as described in Section~\ref{ssec:roi}, we extract 256$\times$256 patches within ROIs with stride 128 pixels. We label a patch as tumor if over 75\% pixels in the patch are annotated as a tumor. Our metastasis detection module is based on the well-known CNN architecture ResNet101~\cite{he2016deep} for patch classification to discriminate between tumor and non-tumor patches.

Although the proposed method seems straightforward, we need to effectively handle gigapixel WSIs to integrate CNN into pN-stage prediction framework for the robust performance, as described below.

\subsubsection{Balanced Patch Sampling}

The areas corresponding to tumor regions often covered only a minor proportion of the total slide area, contributing to a large patch-level imbalance. To deal with this imbalance, we followed similar patch sampling approach used in \cite{liu2017detecting}. In detail, we sample the same number of tumor/normal patches where patches are sampled from each slide with uniform distribution. 

\subsubsection{Patch Augmentation}

\begin{wraptable}{r}{0.45\textwidth}
\vspace{-1.2 cm}
\caption{Patch augmentation details.}
\vspace{+0.1 cm}
\centering
\begin{adjustbox}{max width=0.45\textwidth}
\begin{tabular}{ |c|c| }
 \hline
 Methods & Details \\ 
 \hline
 Translation & random x, y offset in [-8, 8] \\
 Left/right flip & with 0.5 probability \\ 
 Rotation & random angle in [0, 360) \\ 
 \hline
\end{tabular}
\end{adjustbox}
\vspace{-0.6 cm}
\label{table1}
\end{wraptable}

There are only 400 WSIs in Camelyon16 dataset and 500 WSIs in Camelyon17 \texttt{train} set. Patches sampled from same WSI exhibit similar data property, which is prone to overfitting. We perform extensive data augmentation at the training step to overcome small number of WSIs. Since the classes of histopathology image exhibit rotational symmetry, we include patch augmentation by randomly rotating over angles between 0 and 360, and random left-right flipping. Details are shown in Table~\ref{table1}.

\subsubsection{Stain Color Augmentation}

\begin{wraptable}{r}{0.49\textwidth}
\vspace{-0.6 cm}
\caption{Stain color augmentation details.}
\vspace{+0.1 cm}
\centering
\begin{adjustbox}{max width=0.49\textwidth}
\begin{tabular}{ |c|c| }
 \hline
 Methods & Details \\ 
 \hline
 Hue & random delta in [-0.04, 0.04] \\
 Saturation & random saturation factor in [0.75, 1.25] \\
 Brightness & random delta in [-0.25, 0.25] \\
 Contrast & random contrast factor in [0.25, 1.75] \\
 \hline
\end{tabular}
\end{adjustbox}
\vspace{-0.6 cm}
\label{table2}
\end{wraptable}

To combat the variety of hematoxylin and eosin (H\&E) stained color because of chemical preparation difference per slide, extensive color augmentation is performed by applying random hue, saturation, brightness, and contrast as described in Table~\ref{table2}. CNN model becomes robust against stain color variety by applying stain color augmentation at the training step.

\subsubsection{2-Stage Fine-Tuning}

Camelyon16 and Camelyon17 dataset are collected from different medical centers. Each center may use different slide scanners, different scanning settings, difference tissue staining conditions. We handle this multi-center variation by applying the 2-stage fine-tuning strategy. First, we fine-tune CNN with the union set of Camelyon16 and Camelyon17 and then fine-tune CNN again with only Camelyon17 set. The fine-tuned model becomes robust against multi-center variation between Camelyon16 and Camelyon17 set.

\begin{wrapfigure}{r}{0.54\textwidth}
\vspace{-1.4 cm}
 \center
 \includegraphics[width=0.55\textwidth]{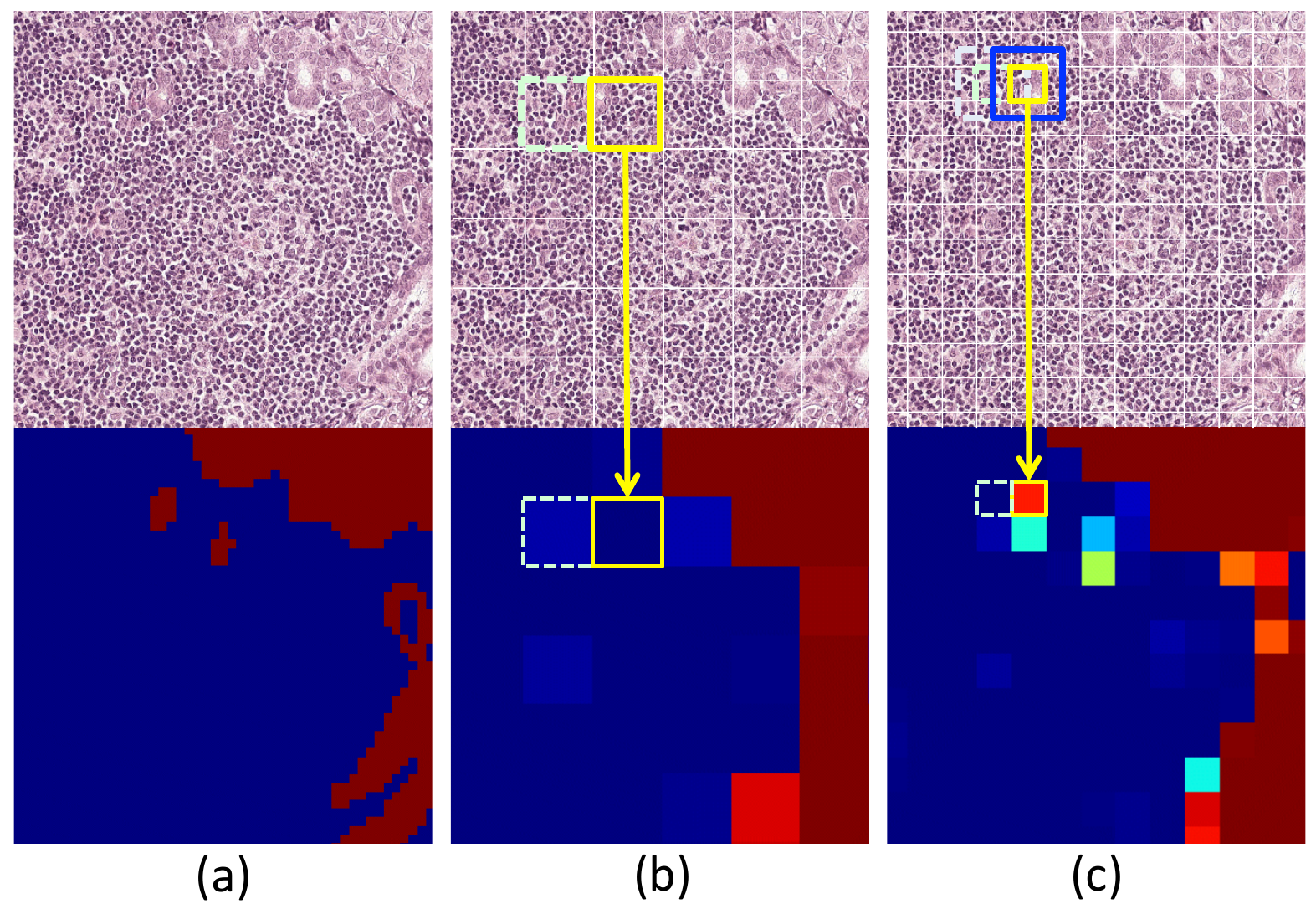}
 \caption{Tiling strategy for dense heatmap. (a) A ground truth; (b) Straightforward tiling strategy; (c) Overlap-tile strategy.}
 \label{fig2}
 \vspace{-0.7 cm}
\end{wrapfigure}

\subsubsection{Overlap Tiling Strategy}
\label{sssec:overlap}

In the prediction stage, probability heatmap is generated by the trained CNN based metastasis detector. A straightforward way to generate a heatmap from WSI is separating WSI into patch size tiles and merging patch level predictions from each tile. However, this simple strategy provides insufficient performance. Instead, we use similar overlap-tile strategy~\cite{ronneberger2015u} for dense heatmap from tiled WSI. As shown in Fig.~\ref{fig2}, the probability heatmap generated by overlap-tile strategy provides denser heatmap than straightforward tiling strategy even though the same classifier is used. By default, we used 50\% overlapped tiles shown in Fig.~\ref{fig2}(c).

\subsection{Lymph Node Classification}
\label{ssec:lymph}

To determine each patient's pN-stage, multiple lymph node slides should be classified into four classes {\small(Normal, Isolated tumor cells (ITC), Micro, Macro)}. For each lymph node WSI, we obtain the 128$\times$ down-sampled tumor probability heatmap through the CNN based metastasis detector (Section~\ref{ssec:cnn}). Each heatmap is converted into a feature vector which is used to build a slide level lymph node classifier. We define 11 types of features based on the morphological and geometrical information. By using converted features, random forest classifier~\cite{breiman2001random} is trained to automatically classify the lymph node into four classes. Finally, each patient's pN-stage is determined by aggregating all lymph node predictions with the given rule~\cite{camelyon17}. We followed the Camelyon17's simplified version of the pN-staging system {\small(pN0, pN0(i+), pN1mi, pN1, pN2)}~\cite{camelyon17}.

\section{Experiments}
\label{sec:experiments}

\subsection{Dataset}
\label{ssec:dataset}

We evaluate our framework on Camelyon16~\cite{bejnordi2017diagnostic} and Camelyon17~\cite{camelyon17} dataset. The Camelyon16 dataset contains 400 WSIs with region annotations for all its metastasis slides. The Camelyon17 dataset contains 1000 WSIs with 5 slides per patient: 500 slides for the \texttt{train} set, 500 slides for the \texttt{test} set. The \texttt{train} set consists of the slide level metastasis annotation. There are 3 categories of lymph node metastasis: Macro {\small(Metastases greater than 2.0 mm)}, Micro {\small(metastasis greater than 0.2 mm or more than 200 cells, but smaller than 2.0 mm)}, and ITC {\small(single tumor cells or a cluster of tumor cells smaller than 0.2mm or less than 200 cells)}.

\begin{wraptable}{r}{0.59\textwidth}
\centering
\vspace{-0.8 cm}
\caption{Details of our Camelyon17 dataset split.}
\begin{adjustbox}{max width=0.59\textwidth}
\begin{tabular}{ |c|c|c|c|c|c|c| }
 \hline
 \multirow{2}{*}{Dataset} & \multicolumn{6}{c|}{\# of patients per each pN-stage} \\ 
 & \multicolumn{1}{c}{pN0} & \multicolumn{1}{c}{pN0(i+)} & \multicolumn{1}{c}{pN1mi} & \multicolumn{1}{c}{pN1} & \multicolumn{1}{c}{pN2} & \multicolumn{1}{c|}{Total} \\
 \hline
 Camelyon17 \texttt{train-M} & 0 & 9 & 11 & 14 & 9 & 43 \\ 
 Camelyon17 \texttt{train-L} & 24 & 3 & 9 & 11 & 10 & 57 \\
 \hline
 \noalign{\smallskip}
 \hline
 \multirow{2}{*}{Dataset} & \multicolumn{6}{c|}{\# of patients per each medical center} \\ 
 & \multicolumn{1}{c}{\enspace Center1\enspace} & \multicolumn{1}{c}{\enspace Center2\enspace} & \multicolumn{1}{c}{\enspace Center3\enspace} & \multicolumn{1}{c}{\enspace Center4\enspace} & \multicolumn{1}{c}{\enspace Center5\enspace} & \multicolumn{1}{c|}{\enspace Total\enspace} \\
 \hline
 Camelyon17 \texttt{train-M} & 7 & 8 & 9 & 10 & 9 & 43 \\ 
 Camelyon17 \texttt{train-L} & 13 & 12 & 11 & 10 & 11 & 57 \\
 \hline
 \noalign{\smallskip}
 \hline
 \multirow{2}{*}{Dataset} & \multicolumn{6}{c|}{\# of WSIs per each metastasis type} \\ 
 & \multicolumn{1}{c}{Negative} & \multicolumn{1}{c}{ITC} & \multicolumn{1}{c}{Micro} & \multicolumn{1}{c}{Macro} & \multicolumn{2}{c|}{Total} \\
 \hline
 Camelyon17 \texttt{train-M} & 110 & 26 & 35 & 44 & \multicolumn{2}{c|}{215} \\
 Camelyon17 \texttt{train-L} & 203 & 9 & 29 & 44 & \multicolumn{2}{c|}{285} \\
 \hline
\end{tabular}
\end{adjustbox}
\vspace{-0.5 cm}
\label{table3}
\end{wraptable}

Since the Camelyon17 set provides only 50 slides with lesion-level annotations in \texttt{train} set, we split 100 patients (total 500 WSIs since each patient provides 5 WSIs) into 43 patients for the Camelyon17 \texttt{train-M} set to train metastasis detection module, 57 patients for the Camelyon17 \texttt{train-L} set to train lymph node classification module. In detail, if patient's any slide include lesion-level annotation, we allocate that patient as a Camelyon17 \texttt{train-M} set. Other patients are allocated as a Camelyon17 \texttt{train-L} set. As shown in Table~\ref{table3}, our split strategy separates similar data distribution between them in terms of the medical centers and metastasis types. 

\subsection{Evaluation Metrics}
\label{ssec:evaluation_metrics}

\subsubsection{Metastasis Detection Evaluation}
\label{ssec:metastasis_detection_module_evaluation}

We used the Camelyon16 evaluation metric~\cite{bejnordi2017diagnostic} on the Camelyon16 dataset to validate metastasis detection module performance. Camelyon16 evaluation metric consists of two metrics, the area under receiver operating characteristic (AUC) to evaluate the slide-level classification and the FROC to evaluate the lesion-level detection and localization. 

\vspace{-0.3 cm}
\subsubsection{pN-stage Classification Evaluation}

To evaluate pN-stage classification, we used the Camelyon17 evaluation metric~\cite{camelyon17}, patient-level five-class quadratic weighted kappa where the classes are the pN-stages. Slide-level lymph node classification accuracy is also measured to validate lymph node classification module performance.

\subsection{Experimental Details}
\label{ssec:setup}

\subsubsection{ROI Extraction Module}
\label{ssec:roi_extraction_module}
For the type of ROI extraction between Otsu threshold and gray value threshold, we determined to use gray value threshold method which is obtained a better performance on Camelyon16 \texttt{train} set. In detail, we convert RGB to gray from 32$\times$ down-sampled WSI and then extract tissue regions by thresholding gray value $>$ 0.8.

\subsubsection{Metastasis Detection Module}
\label{ssec:metastasis_detection_module}

\begin{wraptable}{r}{0.49\textwidth}
\vspace{-1.1 cm}
\caption{Number of training WSIs for metastasis detection module.}
\vspace{+0.1 cm}
\begin{adjustbox}{max width=0.49\textwidth}
\begin{tabular}{ |c|c|c| }
 \hline
 Training data & \# of tumor slides & \# of normal slides \\ 
 \hline
 Camelyon16 \texttt{train} & 110 & 160 \\ 
 Camelyon16 \texttt{test} & 50 & 80 \\ 
 Camelyon17 \texttt{train-M} & 50* & 110 \\
 \hline
 \multicolumn{3}{l}{* only 50 slides include region annotations from total} \\
 \multicolumn{3}{l}{105 tumor slides in Camelyon17 \texttt{train-M} set}
\end{tabular}
\end{adjustbox}
\vspace{-0.6 cm}
\label{table4}
\end{wraptable}

During training and inference, we extracted 256$\times$256 patches from WSIs at the highest magnification level of \SI{0.243}{\micro\metre}/pixel resolution. For training of the patch-level CNN based classifier, 400 WSIs from Camelyon16 dataset and 160 WSIs from Camelyon17 \texttt{train} set are used as shown in Table~\ref{table4}. Total 1,430K tumor patches and 43,700K normal patches are extracted. 

We trained ResNet101~\cite{he2016deep} with initial parameters from ImageNet pretrained model to speed up convergence. We updated batch normalization parameters during fine-tuning because of the data distribution difference between the ImageNet dataset and the Camelyon dataset. We used the Adam optimization method with a learning rate 1e-4. The network was trained for approximately 2 epoch (500K iteration) with a batch size 32 per GPU.

To find hyperparameters and validate performance, we split Camelyon16 \texttt{train} set into our train/val set, 80\% for train and 20\% for validation. For AUC evaluation, we used maximum confidence probability in WSI. For FROC evaluation, we followed connected component approach~\cite{wang2016deep} which find connected components and then report maximum confidence probability's location within the component. After hyperparameter tuning, we finally train CNN with all given training dataset in Table~\ref{table4}.

\begin{table}
\vspace{-0.4 cm}
\centering
\caption{Feature components for predicting lymph node metastasis type.}
\begin{adjustbox}{max width=1.0\textwidth}
\begin{tabular}{ |c|l|c|l| }
 \hline
 No. & Feature description & No. & Feature description\\
 \hline
 1 & largest region's major axis length & 7 & maximum confidence probability in WSI \\
 2 & largest region's maximum confidence probability & 8 & average of all confidence probability in WSI \\
 3 & largest region's average confidence probability & 9 & number of regions in WSI \\
 4 & largest region's area & 10 & sum of all foreground area in WSI \\
 5 & average of all region's averaged confidence probability & 11 & foreground and background area ratio in WSI \\
 6 & sum of all region's area & & \\
 \hline
\end{tabular}
\end{adjustbox}
\label{table5}
\vspace{-1.0 cm}
\end{table}

\subsubsection{Lymph Node Classification Module}
\label{ssec:lymph_node_classification_module}

We generated the tumor probability heatmap from WSI using the metastasis detection module. For the post-processing, we thresholded the heatmap with a threshold of $t=0.9$. We found hyperparameters and feature designs for random forest classifier in Camelyon17 \texttt{train-L} set with 5-fold cross-validation setting. Finally, we extracted 11 features described in Table~\ref{table5}. We built a random forest classifier to discriminate lymph node classes using extracted features. Each patient's pN-stage was determined by the given rule~\cite{camelyon17} with the 5 lymph node slide prediction result.

\subsection{Results}
\label{ssec:results}

\begin{wraptable}{r}{0.45\textwidth}
\vspace{-1.2 cm}
\caption{Metastasis detection results on Camelyon16 \texttt{test} set}
\vspace{+0.1 cm}
\begin{adjustbox}{max width=0.45\textwidth}
\begin{tabular}{ |l|c|c|c| }
 \hline
 \multicolumn{1}{|c|}{Method} & \makecell{Ense-\\mble} & \multicolumn{1}{c|}{AUC} & \multicolumn{1}{c|}{FROC} \\ 
 \hline
 \textbf{Lunit Inc.} & & 0.985 & 0.855 \\
 Y. Liu et al. ensmeble-of-3~\cite{liu2017detecting} & \checkmark & 0.977 & 0.885 \\
 Y. Liu et al. 40X~\cite{liu2017detecting} & & 0.967 & 0.873 \\
 Harvard \& MIT~\cite{wang2016deep} & \checkmark & 0.994 & 0.807 \\ 
 Pathologist*~\cite{bejnordi2017diagnostic} & - & 0.966 & 0.724 \\ 
 \hline
 \multicolumn{4}{l}{* expert pathologist who assessed without a time} \\
 \multicolumn{4}{l}{\enspace constraint} \\
\end{tabular}
\end{adjustbox}
\vspace{-0.8 cm}
\label{table6}
\end{wraptable}

\subsubsection{Metastasis Detection on Camelyon16}
\label{ssec:metastasis_detection_on_camelyon16}

We validated our metastasis detection module on the Camelyon16 dataset. For the fair comparison with the state-of-the-art methods, our model is trained on the 270 WSIs from Camelyon16 \texttt{train} set and evaluated on the 130 WSIs from Camelyon16 \texttt{test} set using the same evaluation metrics provided by the Camelyon16 challenge. Table~\ref{table6} summarizes slide-level AUC and lesion-level FROC comparisons with the best previous methods. Our metastasis detection module achieved highly competitive AUC (0.9853) and FROC (0.8552) without bells and whistles.

\begin{table}
\vspace{-0.3 cm}
\centering
\caption{Top-10 pN-stage classification result on the Camelyon17 leaderboard~\cite{camelyon17}. The kappa score is evaluated by the Camelyon17 organizers. Accessed: 2018-03-02.}

\begin{adjustbox}{max width=1.0\textwidth}
\begin{tabular}{ |l|l|c| }
 \hline
 Team & Affiliation & \makecell{Kappa\\score}\\
 \hline
 \textbf{Lunit Inc.}* & \textbf{Lunit Inc.} & \textbf{0.9203} \\
 HMS-MGH-CCDS & Harvard Medical School, Mass. General Hospital, Center for Clinical Data Science & 0.8958 \\
 DeepBio* & Deep Bio Inc. & 0.8794 \\
 VCA-TUe & Electrical Engineering Department, Eindhoven University of Technology & 0.8786 \\
 JD* & JD.com Inc. - PCL Laboratory & 0.8722 \\
 MIL-GPAT & The Univercity of Tokyo, Tokyo Medical and Dental University & 0.8705 \\
 Indica Labs & Indica Labs & 0.8666 \\
 chengshenghua* & Huazhong University of Science and Technology, Britton Chance Center for Biomedical Photonics & 0.8638 \\
 Mechanomind* & Mechanomind & 0.8597 \\
 DTU & Technical University of Denmark & 0.8244 \\
 \hline
 \multicolumn{3}{l}{* Submitted result after reopening the challenge} \\
\end{tabular}
\end{adjustbox}
\label{table7}
\vspace{-1.2 cm}
\end{table}

\subsubsection{pN-stage Classification on Camelyon17}

For validation, we first evaluated our framework on Camelyon17 \texttt{train-L} set with 5-fold cross-validation setting. Our framework achieved 0.9351 slide-level lymph node classification accuracy and 0.9017 patient-level kappa score using single CNN model in metastasis detection module. We trained additional CNN models with different model hyperparameters and fine-tuning setting. Finally, three model was ensembled by averaging probability heatmap and reached 0.9390 slide-level accuracy and 0.9455 patient-level kappa score with the 5-fold cross-validation.

Next, we evaluated our framework on the Camelyon17 \texttt{test} set and the kappa score has reached 0.9203. As shown in Table~\ref{table7}, our proposed framework significantly outperformed the state-of-the-art approaches by large-margins where it achieves better performance than the previous winning method (HMS-MGH-CCDS) of the Camelyon17 challenge. 

\begin{table}
\centering
\caption{Slide-level lymph node classification confusion matrix comparison on the Camelyon17 \texttt{test} set. The confusion matrix is generated by the Camelyon17 organizers.}
\begin{adjustbox}{max width=0.95\textwidth}
\parbox{.48\textwidth}{
\centering
\begin{adjustbox}{max width=0.45\textwidth}
\begin{tabular}{|r|rrrr|}
\hline
\multirow{2}{*}{} & \multicolumn{4}{c|}{Predicted} \\
& \multicolumn{1}{c}{Negative} & \multicolumn{1}{c}{ITC} & \multicolumn{1}{c}{Micro} & \multicolumn{1}{c|}{Macro} \\
\hline
\parbox[t]{2mm}{\multirow{4}{*}{\rotatebox[origin=c]{90}{Reference}}} 
Negative & \cellcolor{Gray}\textbf{96.15\%} & 3.08\% & 0.77\% & 0.00\% \\
ITC & \textcolor{red}{55.88\%} & \cellcolor{Gray}\textbf{11.76\%} & 32.35\% & 0.00\% \\
Micro & \textcolor{red}{9.64\%} & \textcolor{red}{2.41\%} & \cellcolor{Gray}\textbf{85.54\%} & 2.41\% \\
Macro & \textcolor{red}{3.25\%} & \textcolor{red}{0.00\%} & \textcolor{red}{5.69\%} & \cellcolor{Gray}\textbf{91.06\%} \\
\hline
\multicolumn{5}{c}{(a) \textbf{Ours}}
\end{tabular}
\end{adjustbox}
}
\hfill
\parbox{.48\textwidth}{
\centering
\begin{adjustbox}{max width=0.45\textwidth}
\begin{tabular}{|r|rrrr|}
\hline
\multirow{2}{*}{} & \multicolumn{4}{c|}{Predicted} \\ 
& \multicolumn{1}{c}{Negative} & \multicolumn{1}{c}{ITC} & \multicolumn{1}{c}{Micro} & \multicolumn{1}{c|}{Macro} \\
\hline
\parbox[t]{2mm}{\multirow{4}{*}{\rotatebox[origin=c]{90}{Reference}}} 
Negative & \cellcolor{Gray}\textbf{95.38\%} & 0.38\% & 4.23\% & 0.00\% \\
ITC & \textcolor{red}{76.47\%} & \cellcolor{Gray}\textbf{14.71\%} & 8.82\% & 0.00\% \\
Micro & \textcolor{red}{13.25\%} & \textcolor{red}{1.20\%} & \cellcolor{Gray}\textbf{78.31\%} & 7.23\% \\
Macro & \textcolor{red}{1.63\%} & \textcolor{red}{0.00\%} & \textcolor{red}{12.20\%} & \cellcolor{Gray}\textbf{86.18\%} \\
\hline
\multicolumn{5}{c}{(b) HMS-MGH-CCDS}
\end{tabular}
\end{adjustbox}
}
\end{adjustbox}
\label{table8}
\vspace{-0.5 cm}
\end{table}

Furthermore, the accuracy of our algorithm not only exceeded that of current leading approaches (bold black color in Table~\ref{table8}) but also significantly reduced false-negative results (red color in Table~\ref{table8}). This is remarkable from a clinical perspective, as false-negative results are most critical, likely to affect patient survival due to consequent delay in diagnosis and appropriate timely treatment.

\vspace{-0.2 cm}
\section{Conclusion}
\label{sec:conclusion}

We have introduced a robust and effective method to predict pN-stage from lymph node histological slides, using CNN based metastasis detection and random forest based lymph node classification. Our proposed method achieved the state-of-the-art result on the Camelyon17 dataset. In future work, we would like to build an end-to-end learning framework for pN-stage prediction from WSIs.

%
%

\bibliographystyle{splncs}
\bibliography{refs}

\end{document}